\documentclass{article}

\PassOptionsToPackage{numbers, compress}{natbib}

\usepackage[final]{nips_2018}
\usepackage{graphicx}
\usepackage{subcaption}
\usepackage{array}

\usepackage[utf8]{inputenc} 
\usepackage[T1]{fontenc}    
\usepackage{hyperref}       
\usepackage{url}            
\usepackage{booktabs}       
\usepackage{amsfonts}       
\usepackage{nicefrac}       
\usepackage{microtype}      
\usepackage{color}

\title{Interpretable deep learning for guided structure-property explorations in photovoltaics}

\author{
  Balaji Sesha Sarath Pokuri \\
  Department of Mechanical Engineering\\
  Iowa State University\\
  Ames, IA 50011 \\
  \texttt{balajip@iastate.edu} \\
  \And
  Sambuddha Ghosal
  \\
  Department of Mechanical Engineering, \\ Department of Computer Science \\
  Iowa State University \\
  Ames, IA 50011 \\
  \texttt{sghosal@iastate.edu} \\
    \And
    Apurva Kokate\\
  Department of Computer Science, \\
  Iowa State University \\
  Ames, IA 50011 \\
  \texttt{akokate@iastate.edu} \\
  \And
  Baskar Ganapathysubramanian \\
  Department of Mechanical Engineering\\
  Iowa State University\\
  Ames, IA 50011 \\
  \texttt{baskarg@iastate.edu} \\
  \And
  Soumik Sarkar \\
  Department of Mechanical Engineering\\
  Iowa State University\\
  Ames, IA 50011 \\
  \texttt{soumiks@iastate.edu} \\
}

\begin{document}

\maketitle

\begin{abstract}

The performance of an organic photovoltaic device is intricately connected to its active layer morphology. This connection between the active layer and device performance is very expensive to evaluate, either experimentally or computationally. Hence, designing morphologies to achieve higher performances is non-trivial and often intractable.
To solve this, we first introduce a deep convolutional neural network (CNN) architecture that can serve as a fast and robust surrogate for the complex structure-property map. Several tests were performed to gain \emph{trust} in this trained model. Then, we utilize this fast framework to perform robust microstructural design to enhance device performance.
\end{abstract}

\section{Introduction}

Mapping microstructure to macro-scale property of materials and controlling such relationships has been an important theme of modern materials research~\cite{ju2017designing}. Despite the investment of millions of research hours and dollars~\cite{dmref}, this largely remains elusive to the research community. A major reason is that the map from microstructure (domain)	to property (co-domain) is inherently non-surjective and highly non-linear, making incremental judgments based on incremental modifications infeasible. Added to this is the infinite dimensionality of microstructure space, which makes the description of the domain in-exhaustible. Hence, estimating the property of one morphology from another visually similar morphology is not trivial -- each morphology will require a full scale experiment/simulation for reliable quantification.
In this paper, we discuss one such problem and present a solution using modern
machine learning paradigms. More specifically, we look at solving the
relationship between active layer microstructure and performance of an Organic
Solar Cell (OSC). OSCs utilize a bulk heterojunction morphology in the active layer
to efficiently convert incident solar radiation into electrical energy~\cite{ray2012can}. The total power conversion efficiency is very intricately related to this material distribution
if acceptor and donor polymers in the active layer (See Sec.~\ref{sec:prob}).
This relationship has several contradicting and competing factors. For example,
larger interfacial area between acceptor and donor regions enables higher production
of charges, but simultaneously also increases the loss of charges through
recombination. While larger domains can help transport charges faster,
they also impede excitons from dissociating into charges. These contradictory
features make this an exciting area of research, finally culminating in the search for the most efficient microstructures for solar energy conversion.

Traditionally, this structure-property map was computationally resolved either through a
detailed and tedious PDE simulation~\cite{kodali2012computer} or Kinetic Monte Carlo
simulation~\cite{casalegno2010methodological,meng2009dynamic,marsh2007microscopic,watkins2005dynamical} or through identification of surrogate descriptors of performance~\cite{wodo2012graph}. While the first method is robust and accurate, it
is limited by the size of microstructure. Often it requires massive, robust, and well established computational resources along with a perfectly scaling computational model
to solve for even simple representations of morphologies. Hence such analysis was often limited to 2D microstructure quantification, although 3D quantification is not uncommon~\cite{kodali2012computer}. The second method overcomes most of these computational challenges. Typically intuitive features of the microstructure such as domain sizes and orientations, connectivities and islands are quantified (using graphs) instead of performing a full scale PDE simulation. This technique is intuitive and requires substantially less computational resources (although memory requirements are higher). Subsequently, it enables simultaneous parallel analysis of multiple morphologies at once. Moreover, close to linear dependence of several stage efficiencies on these microstructural descriptors was also shown~\cite{wodo2015automated}. However, since the dimensionality of the domain is infinite, the intuitive metrics tend to be a limited description of the bigger reality.

\begin{figure}[!ht]
  \centering
  \includegraphics[width=\textwidth]{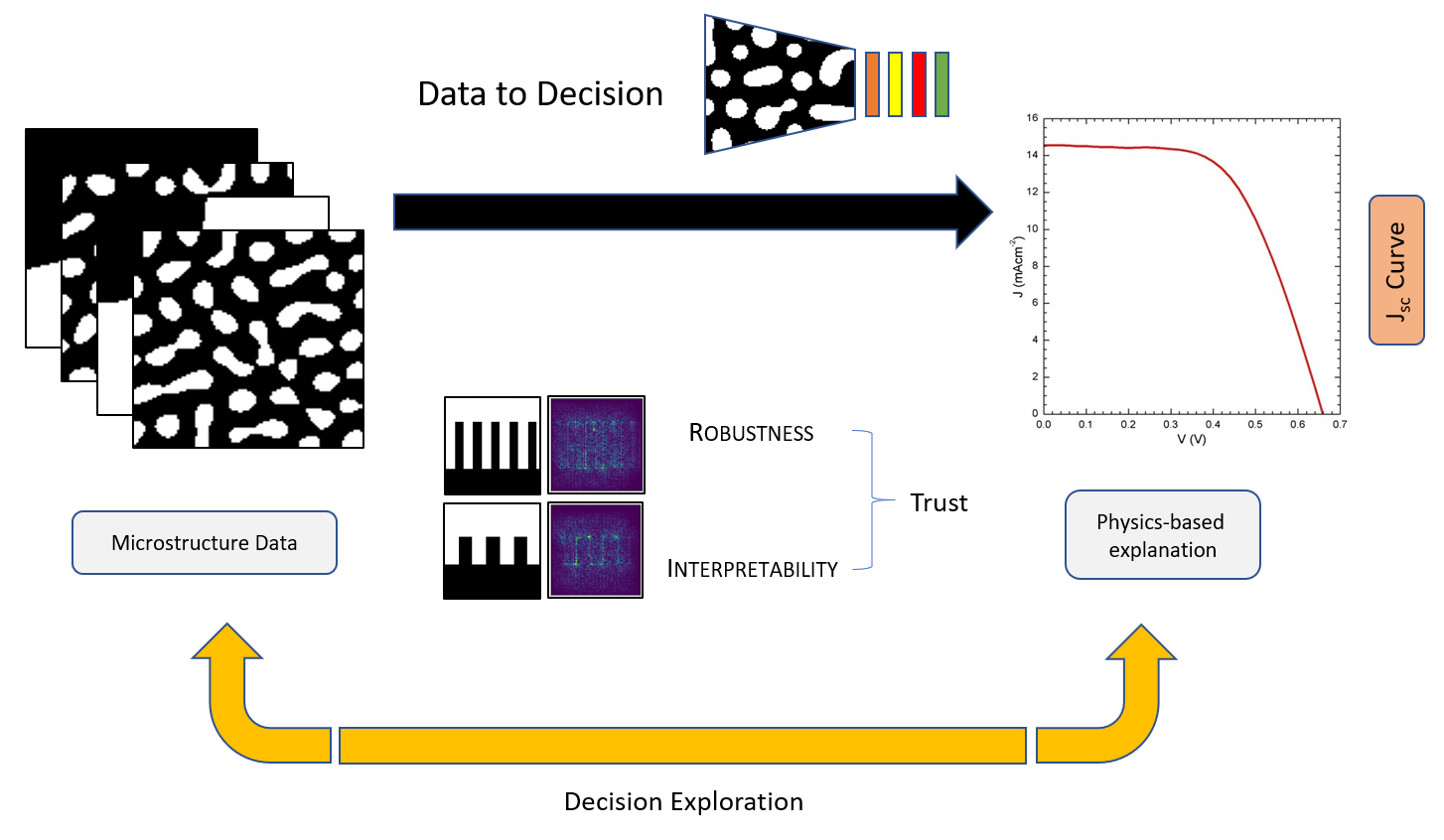}
  \caption{DLSP framework: We construct a forward map from morphology to performance. Upon building \emph{trust} in this trained model, it was further used for performing manual and automated design. }
  \label{fig:flow}
\end{figure}

In this work, we take the approach called DLSP (Deep Learning based Structure-Property exploration) involving machine learning to solve this complex map of microstructure to performance (depicted in Fig.~\ref{fig:flow}). Modern machine learning has been shown to be a powerful tool to learn complex phenomena~\cite{Esteva}, especially image recognition ~\cite{Yamins, Alip}. This technique has the potential to identify features in the microstructure that are not intuitive to the simple eye. By posing the task of enumerating microstructure informed performance as an image recognition problem, we can easily and confidently extend current implementations of neural networks to this problem. Furthermore, interpreting the trained network can give both trust in training as well as insights into the most impactful features of a microstructure, thereby enabling researchers to tailor processes to create such features.

The rest of the paper is organised as follows:
In Sec.~\ref{sec:prob}, we discuss the physics behind the problem of morphology design.
Subsequently, a short discussion on the methods for data generation and labelling
is discussed. The details of network architecture and other machine learning parameters
are discussed in Sec.~\ref{sec:archi}. In Sec.~\ref{sec:results}, the results of
training, validation and testing on in-sample and out-of-sample morphologies and performance
interpretations are presented. The benefits of such fast interpretable forward surrogate
model are demonstrated through the design of property-maximized-microstructure in Sec.~\ref{sec:design}.
Finally in Sec.~\ref{sec:concl}, key takeaways from this work and future directions are discussed.


\section{Physics of structure-property explorations in photovoltaics}
\label{sec:prob}
\subsection{Organic Photovoltaics}

Organic photovoltaic devices are energy harvesting devices which employ organic materials for solar energy conversion. These provide multiple advantages over traditional silicon based cells, like flexibility, transparency and ease of manufacturability. They however are limited by their efficiency of operation. Although major breakthroughs in processing and materials have improved the efficiency drastically, they still lag behind the traditional photovoltaics.

The efficiency of these devices is intricately dependant on the material distribution/morphology in the active layer. The active layer generally is a bulk hetero-junction, enabling multiple sites for charge generation. Several features of the morphology have different roles in the process of converting solar energy. The ability to change these morphological features by changing the processing protocol is a major source of control in these devices.

There are several stages during the solar power conversion in an OPV. Firstly, the incident solar energy generates excitons in the donor phase. These excitons are highly unstable and need to diffuse to a nearest interface with the acceptor material to separate into positive and negative charges. This diffusion to the interface is critical to evaluate the efficiency of absorption of incident light. The dissociation of excitons to form charges depends on the nature of interface and the materials in the interface. For example, interfaces with non-aligned crystal boundaries show lower dissociation than those with aligned crystals . In the next stage, these charges (positive charge in the donor and negative charge in the acceptor) need to be drifted to the respective electrode to produce electricity. Usually, this drift is provided by the potential difference between the two electrodes. However, these charges also encounter other interfaces which have pairs of positive and negative charges, leading to potential recombination. In summary, the total efficiency of the active layer involves exciton production, charge separation and charge transportation efficiencies.

In this context, quantifying the dependence of the device as well as stage efficiencies becomes a critical part in developing strategies to design processing conditions. It can already be seen that the role of morphology cannot be over-estimated in the power conversion efficiency. Hence strategies were developed to quantify the efficiencies these morphologies. 
While these techniques are robust and rigorous, they are expensive and time intensive. This makes them infeasible for further designing morphologies, which often requires several quantifications. So, we turn to modern fast methods of quantifying data, especially images. We represent the morphologies as images and take advantage of the deep convolutional neural networks to do performance based classification.

\begin{figure}[!ht]
  \centering
  \includegraphics[width=\textwidth]{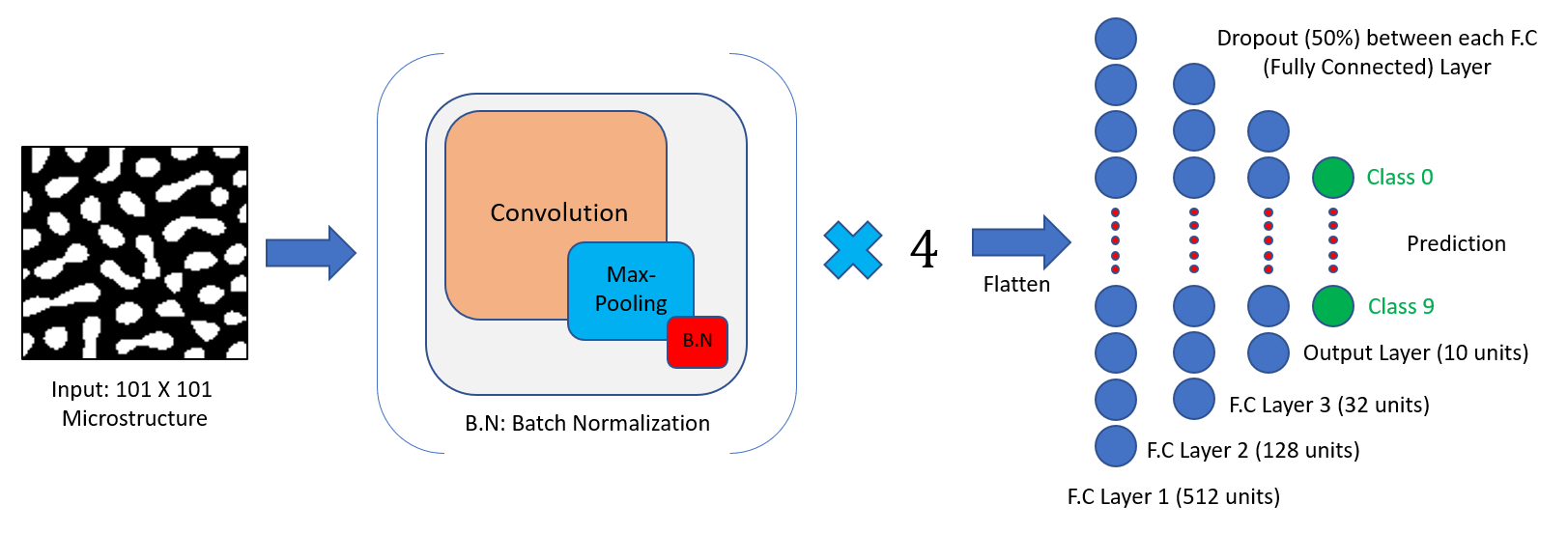}
  \caption{Proposed CNN architecture. Note that this is much shallower and has less trainable parameters compared to VGG-16 and ResNet-50.}
  \label{fig:architecture}
\end{figure}

\subsection{Data generation and quantification}
In order to train the network (to be described in Sec.~\ref{sec:archi}), we generate a dataset of microstructural images using the Cahn-Hilliard equation. This process generates time series of images that can be treated as independent images for the sake of training. This method helps to quickly produce several thousands of images within a very short amount of time. Previous analysis using these images can be found in~\cite{wodo2015automated}. A characteristic of this procedure for generating morphologies through simulation is their similarity to morphologies in real active layers produced during thermal annealing. In the morphology, the domains are similar in size and have smooth interface contours. These characteristics will also help us to build trust in the training process by manually create morphologies breaking these characteristics and testing the performance of the trained network. More implications of this will be in discussed in Sec.~\ref{sec:results}. Using data augmentation techniques over the originally simulated data, we finally produce a dataset of nearly $65,000$ (2D) morphologies. Each morphology is a gray-scale image of size 101px $\times$ 101px.

These morphologies were then characterized using an in-house physics based simulator~\cite{kodali2012computer}. This simulator uses steady state excitonic drift diffusion equation to model the processes of exciton dissociation and charge transport. We use the short circuit current $J_{sc}$ as a means of labelling the data. The whole data was divided into $10$ classes which are equally spaced between the best($J_{sc} = 7mA/cm^2$) and worst performing($J_{sc} = 0.2mA/cm^2$) in the data.

\section{Proposed Deep Learning framework}
\label{sec:archi}

Convolutional Neural Networks (CNN) are well-established architectures for classification of images. They have proved to perform outstandingly well in terms of efficiently extracting complex features from images and function as a classification technique when provided with sufficient data \cite{Esteva,Yamins,Alip,Mnih,Silver,Ghosal4613,ghosalhigh}. Hence, to address the task above, we develop a CNN based architecture to classify the morphologies. The data label, originally a continuous value, is discretized into $10$ bins. The network architecture is depicted in Fig.~\ref{fig:architecture}, with $1.2$ million learning parameters. This architecture was taken from~\cite{bojarski2016end}, where the original aim was to determine steering angles for a car from the dash camera image. The task is similar here also -- both use images from time series data with predefined label to train. Our network is slightly shallower, to enable better training, enhanced with better feature visualization. This is also to prevent problems that can arise with deeper networks -- vanishing (or exploding) gradients \cite{Glorot}, which provide a hindrance to convergence, and the accuracy saturation with increasing depth. Shallower networks also provide better feature visualizations -- in this case, we use saliency maps \cite{Simon} to visualize learnt features(Sec.~\ref{sec:results}). The network was trained for approximately $70$ epochs (with $18s$ per epoch) with a learning rate of $0.0001$, on the $45,000$-image training set to reach the desired accuracy ($95.41\%$). The crossentropy (categorical) loss (or cost) function along with the Adam optimizer \cite{Kingma} was used to minimize the error.

Apart from this, we also tested two standard architectures with our dataset:
\begin{itemize}
    \item VGG-16 (learning parameters ~ $50$ million) , with learning rate of $0.0001$, batch size of $128$ initialized with random weights was also trained on the training dataset, achieving a test accuracy of $96.61\%$ at epoch $70$ (with $180s$ per epoch).

    \item ResNet-$50$ (learning parameters ~ $23$ million) , with learning rate of $0.0001$, batch size of $128$ initialized with random weights was also trained on the training dataset, achieving a test accuracy of $96.45\%$ at epoch $70$ (with $580s$ per epoch).

\end{itemize}

We can see that our custom (shallower) CNN, while with lower training parameters, performs very similar to the established deep CNNs. Therefore, we defer the choice of selecting a "good" network to be based on the learnt features and out-of-sample performance and not just the accuracy/f1-score of the testing dataset.

\section{Results}
\label{sec:results}

\begin{figure}[!ht]
  \centering
  \includegraphics[width=\textwidth]{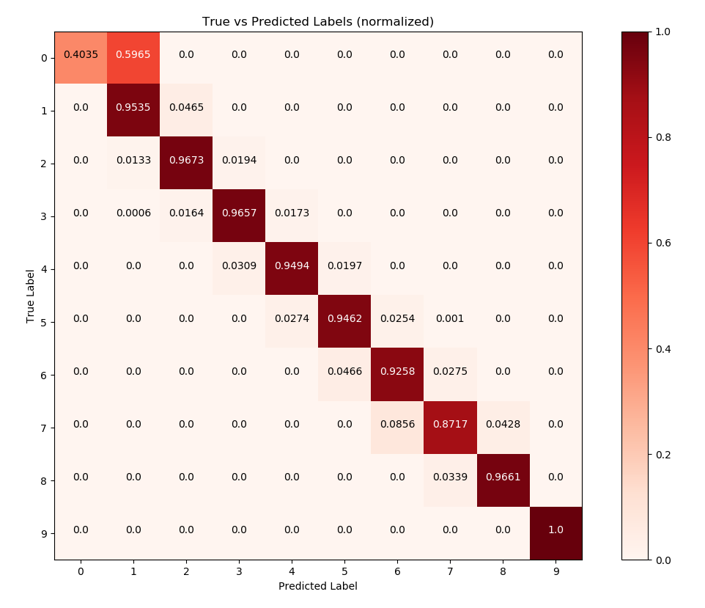}
  \caption{Confusion matrix for insample test predictions. Notice the heavily diagonally dominant matrix, indicating a very good classification accuracy.}
  \label{fig:insample_conf}
\end{figure}

Figure~\ref{fig:insample_conf} shows the confusion matrices for in-sample test data classification. It had an accuracy of $95.41\%$ and F1-score of $~94.45\%$. From the confusion matrix, it can clearly be seen that most of the classification is correct, and those which are wrongly predicted are only off by one class. This wrong prediction is not unexpected, as we are binning a continuous variable into non-overlapping classes. The edge cases have the potential to be misclassified. It is also to be noted that other two standard architectures show similar confusion matrices, with similar prediction accuracies.

\subsection{Out-of-sample testing}

\begin{figure}
    \centering
    \begin{subfigure}{0.28\textwidth}
    \centering
    \includegraphics[width=\textwidth]{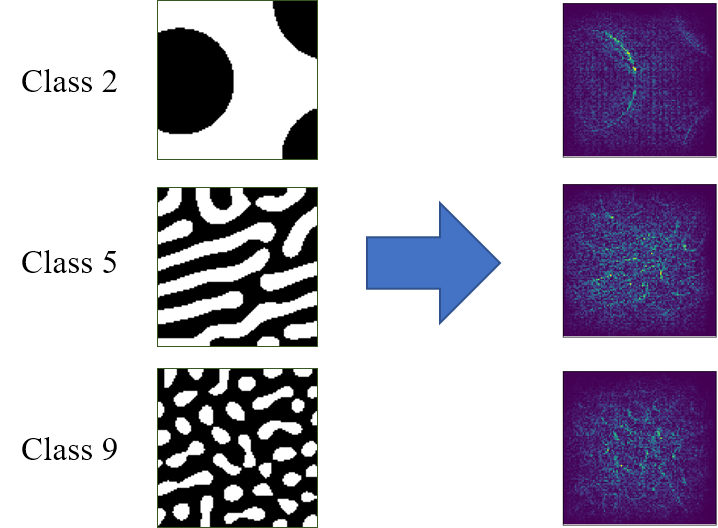}
    \caption{Saliency map for in-sample testing morphologies}
    \label{fig:sal_insample}
    \end{subfigure}
    \begin{subfigure}{0.42\textwidth}
    \centering
    \includegraphics[width=0.8\textwidth]{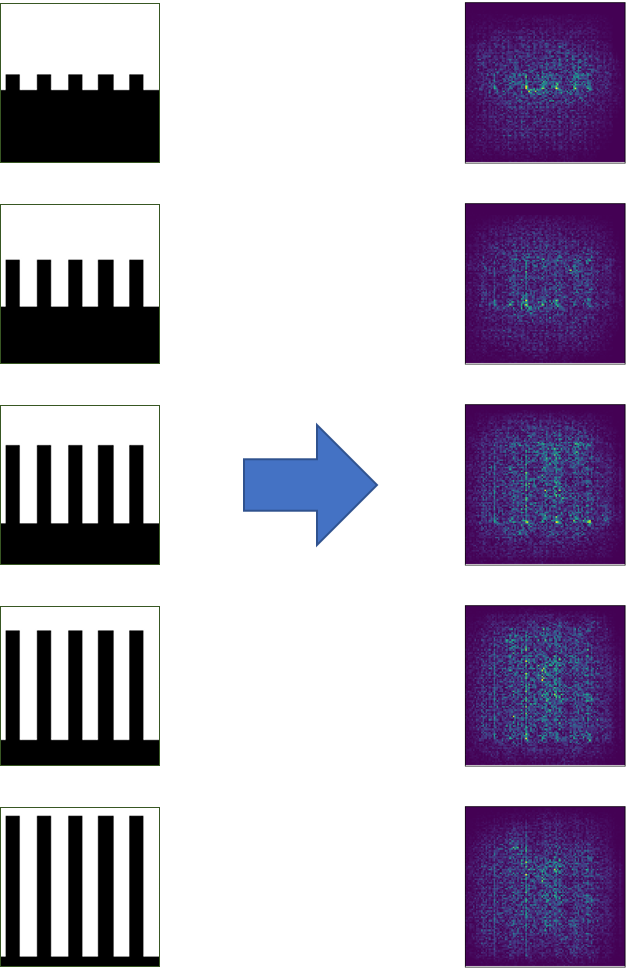}
    \caption{Saliency maps for out-of-sample morphologies}
    \label{fig:sal_outsample}
    \end{subfigure}
    \begin{subfigure}{0.28\textwidth}
    \centering
    \includegraphics[width=\textwidth]{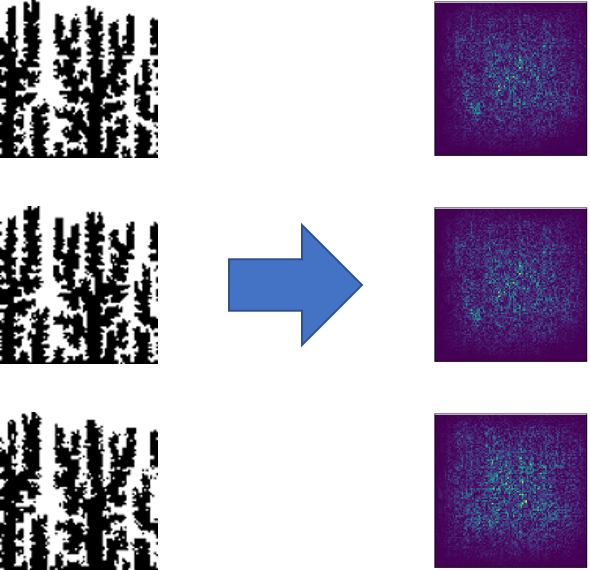}
    \caption{Saliency maps for 'optimized' structures from~\cite{du2018microstructure}}
    \label{fig:sal_omega}
    \end{subfigure}
    \caption{Saliency maps and performance of our custom trained CNN. Note how the saliency maps closely follow the interface regions in the microstructure. It should also be noted that the networks shows good performance even on samples outside the training dataset.}

    \label{fig:sal}
\end{figure}

While it is a known fact in the machine learning community that neural networks in their training phase can possibly overfit (depending on the model capacity, amount of training data and training hyperparameters), we resort to two methods of checking the robustness of our trained network(s). As noted earlier, the morphology data used for training is generated by solving a PDE. This inherits certain properties to the data such as smooth contours and uniform domain sizes. Hence we try to systematically break these assumptions about the dataset and see the performance of the network. Firstly, we test the network on a columnar structure (Fig.~\ref{fig:sal_outsample}). This is a widely studied structure, often considered as a high performing morphology. This is a completely out-of-sample data. This columnar morphology also has several sharp interface contours, which are completely absent in the training dataset. The results are in Fig.~\ref{fig:sal_outsample}. The actual $J_{sc}$ values from a full scale drift-diffusion simulation (along with the corresponding true label) are also presented. We can see how the network is able to predict the correct label corresponding to each of the columnar microstructures.

Next, we also test our framework with the "high" performing morphologies described in~\cite{du2018microstructure}. These morphologies were obtained through optimization strategies using graph based surrogate performance descriptors. We can also see how our network has rightly identified (Fig.~\ref{fig:sal_omega}) \emph{all} these as high performing class label \emph{9}.

\subsection{Interpretability}

We next query the network to understand the learnt features. For this, we use saliency maps~\cite{Simon, montavon2017methods} to identify important features of the image input. Saliency map visualization is a visualization technique that generates heat maps on test images that bring out (highlight) the regions (microstructure regions, for our case) the trained CNN model focuses on to generate it's classification output. The heat-map signifies the aspect that regions highlighted more than the others carry a greater significance towards the model's classification output. Fig.~\ref{fig:sal} shows the saliency maps for morphologies in the data, columnar structures and the "high" performing morphologies identified in~\cite{du2018microstructure}.

We can see, in all Figs.~\ref{fig:sal_insample},~\ref{fig:sal_outsample}~\&~\ref{fig:sal_omega}, how the network highlights interface between the acceptor and donor regions. This is a vital test of the learning capabilities of the network. In terms of the underlying physics, the interface is the most critical feature of performance. Properly configured interface enables charge separation and hence better conversion. Poorly configured interface enables poor dissociation and higher recombination, which depreciates conversion efficiency. Finally, interface further away from the top and bottom electrodes are more critical, as the charges produced at these locations have a higher chance of recombination. We can see from Fig.~\ref{fig:sal} how the network is able to find the right type of interface as critical to device performance.

\begin{figure}[!ht]
  \centering
  \includegraphics[scale=0.35]{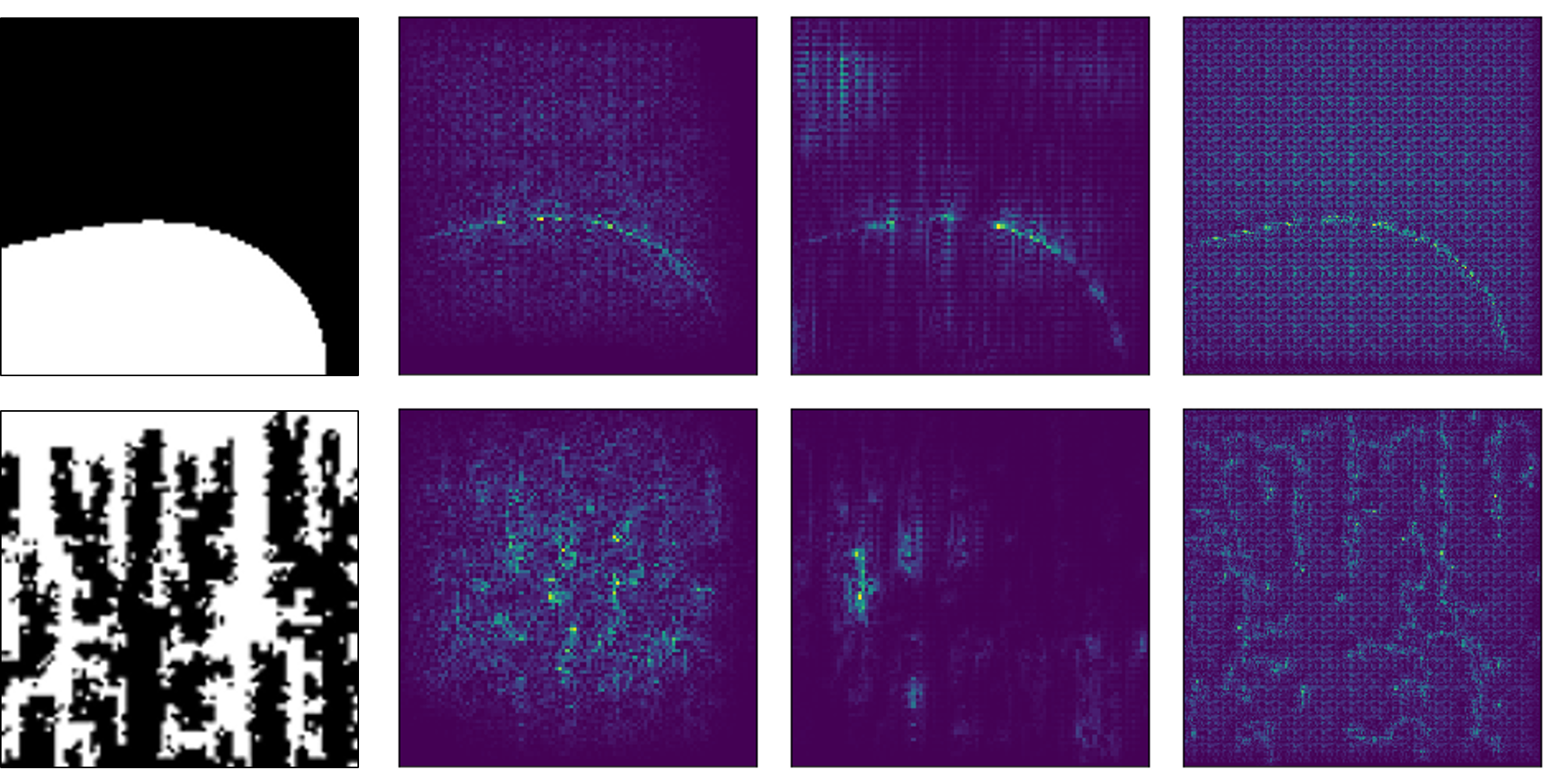}
  \caption{Comparison of Saliency map outputs for our Custom Model (second column), VGG-net (third column) and ResNet-50 (fourth column), with input images shown in the first column: top row shows an example image for class 0, bottom row shows an example image from omega morphologies (correctly predicted as class 9 by our custom model)}
  \label{fig:sal_compare}
\end{figure}

Furthermore, we also observe from Fig.~\ref{fig:sal_compare} that the saliency maps from the standard deep networks (VGG-$16$ and ResNet-$50$) are unable to locate these above features. Although, the test accuracy is slightly higher than our custom network, we see that the saliency outputs do not provide us with any understandable information. This observation is in line with~\cite{Yosinski}, where it was shown that deeper models are harder to explain than their shallower counterparts even though they may achieve a higher classification accuracy. These results signify the importance of tailoring architectures to the application. Thus, for performing morphology design, we use the custom architecture as a surrogate map from the microstructure space to the performance space.

\section{Morphology design}
\label{sec:design}


Having developed a fast and \emph{trust-worthy} surrogate map from the microstructures to the performance, we can use it enable microstructural design. In this section, we show two separate techniques, manual and automated, for designing microstructures. The goal of both these techniques is to explore (uncanny) morphologies that demonstrate superior performance. Traditionally, this was generally achieved through a conventional optimization strategy, like simulated annealing, where an initial morphology is tweaked repeatedly to achieve superior performance. At every stage, the current morphology is evaluated for its performance. Subsequently, the whole process requires several computationally expensive evaluations and hence becomes time consuming. But, with the above demonstrated framework, evaluating the morphology becomes significantly faster and easier. Hence it provides an very powerful way to quickly 'evolve' morphologies to optimize performance.

\subsection{Manual design}
\begin{figure}[!ht]
    \centering
    \includegraphics[width=0.8\textwidth]{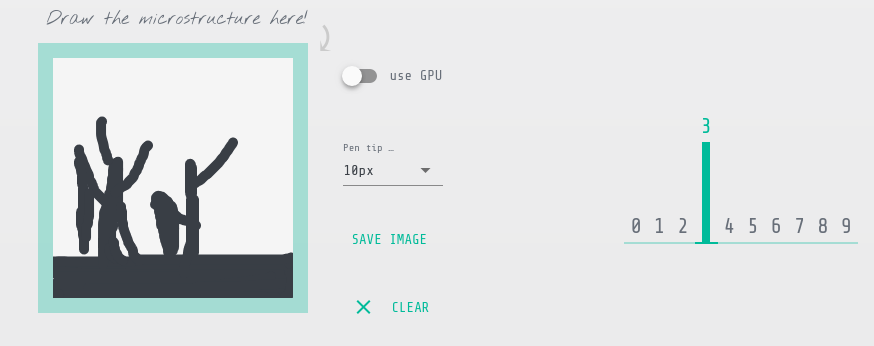}
    \caption{Browser interface for performing manual exploration and design.}
    \label{fig:web_interf}
\end{figure}

\begin{figure}
\begin{subfigure}{0.13\textwidth}
\includegraphics[width=\textwidth, trim={0 0 550px 40px}, clip]{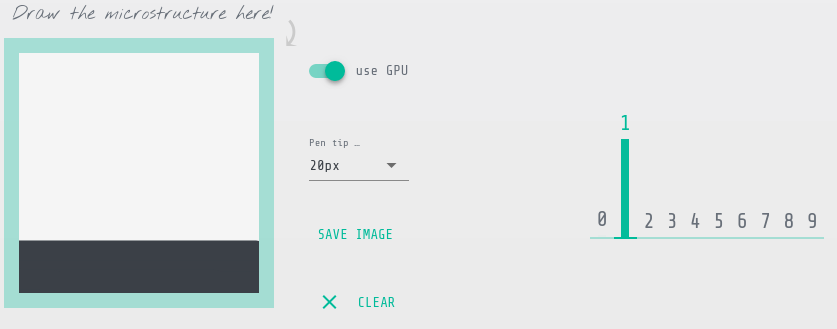}
\caption{Class 1}
\end{subfigure}
\begin{subfigure}{0.13\textwidth}
\includegraphics[width=\textwidth, trim={0 0 550px 40px}, clip]{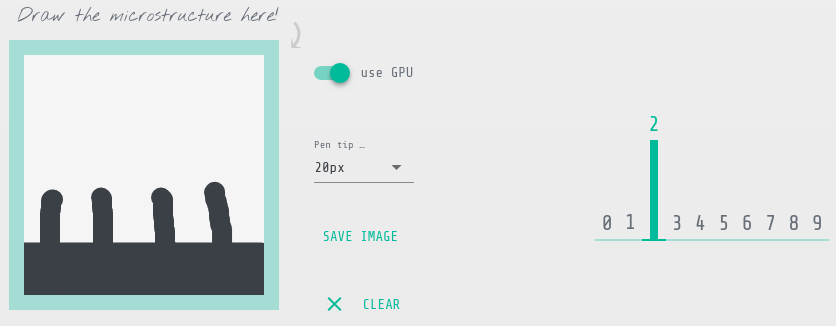}
\caption{Class 2}
\end{subfigure}
\begin{subfigure}{0.13\textwidth}
\includegraphics[width=\textwidth, trim={0 0 550px 30px}, clip]{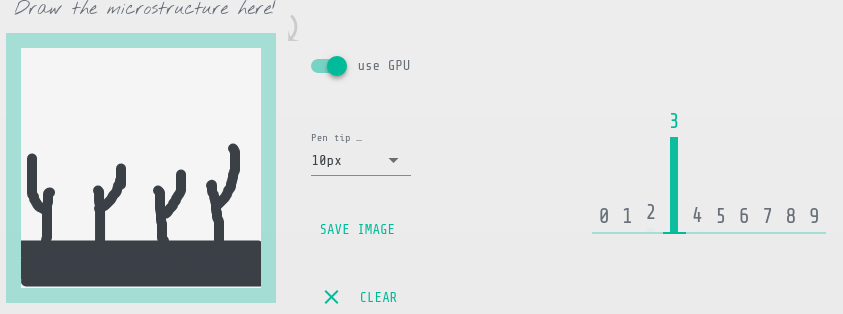}
\caption{Class 3}
\end{subfigure}
\begin{subfigure}{0.13\textwidth}
\includegraphics[width=\textwidth, trim={0 0 550px 40px}, clip]{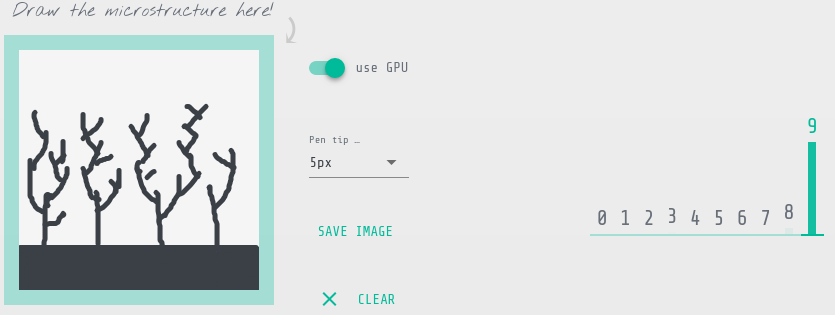}
\caption{Class 9}
\end{subfigure}
\begin{subfigure}{0.13\textwidth}
\includegraphics[width=\textwidth, trim={0 0 550px 40px}, clip]{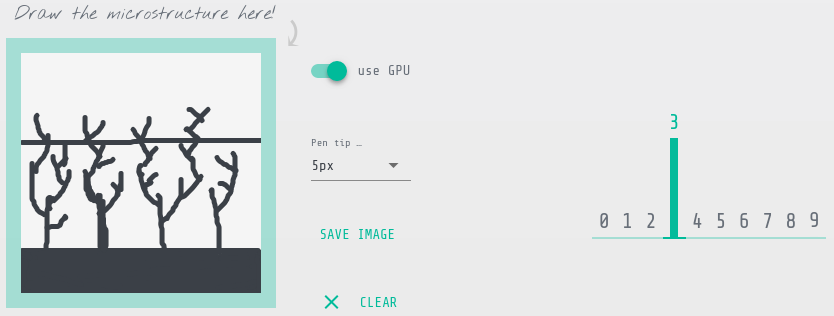}
\caption{Class 3}
\end{subfigure}
\begin{subfigure}{0.13\textwidth}
\includegraphics[width=\textwidth, trim={0 0 550px 40px}, clip]{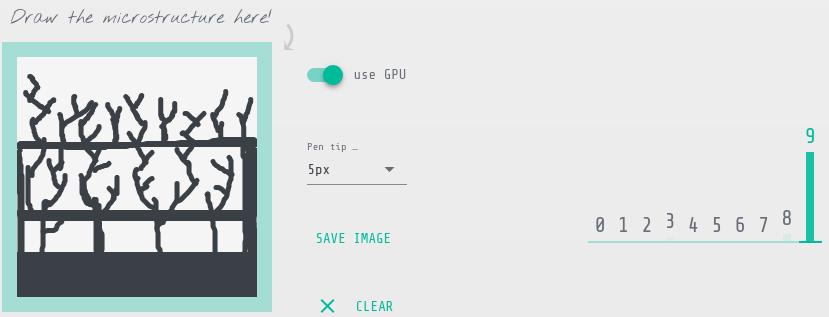}
\caption{Class 9}
\end{subfigure}
\begin{subfigure}{0.13\textwidth}
\includegraphics[width=\textwidth, trim={0 0 550px 40px}, clip]{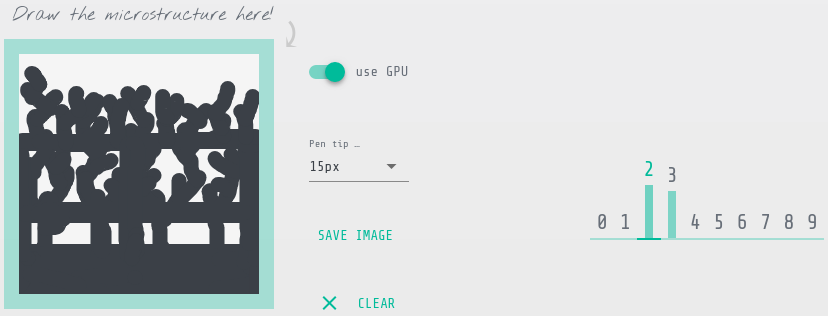}
\caption{Class 2/3}
\end{subfigure}
\caption{Exploration by manual design: Notice how the trained network captures the underlying physics --  addition of non-interfering features improves performance. When a `blocking' layer was added in (e), the performance drops. Also note that increasing domain size (in (g)) leads to performance drop, in conjunction with reduced exciton dissociation efficiency and increased recombination.}
    \label{fig:manual_design}
\end{figure}

In order to enable manual exploration, we created a browser interface (Fig.~\ref{fig:web_interf}) that enables the user to interactively modify morphologies to both visualize and improve morphology performance. Using this interface, the user can incrementally add changes to the initial morphology that can improve the predicted performance. Since the performance assessment is done by the trained CNN, the whole process happens real-time. Fig.~\ref{fig:manual_design} shows how one can modify images to include several features of varying sizes, with the aim of improving performance. This tool can in turn help identify features of morphology that affect the performance.

\subsection{Automated design}
While the above interface may serve very well to understand the influence of morphological features on performance, it is not feasible to perform a full scale morphology design. Manual exploration limits the full exploration of the \emph{best performing morphology manifold}. Thus, to fully explore this space, we link this fast surrogate with a probabilistic optimization algorithm to find those optimal structures. More specifically we use a population based incremental learning (PBIL) approach to model morphologies and evolve them to achieve optimum performance. PBIL estimates the explicit probability distribution of the optimal morphology. The multi variate probability distribution is stored as a probability matrix $P$ of the 2D morphology, i.e., each pixel is associated with a probability. This matrix $P$ is updated as follows: the optimization starts with a given probability matrix, generally based on the intuition of the researcher. Subsequently, $n$ morphology instances are sampled around this matrix $P$. For each realization, the DLSP is deployed to evaluate the performance, $f_j, j\in [1,n]$. Then $n_b$ best samples ($n_b < n$) are used to calculate, $P_u$, the probabilistic update matrix. Next, the probability vector is updated according to $P = P \cdot(1-l_r) + P_b \cdot l_r$, where $l_r$ is the learning rate. Intuitively, the update step reinforces features present in the best performing morphologies, and dampens those missing. The algorithm terminates by standard criteria (iteration limits and improvement bounds). Only the probability matrix is stored and multiple realizations' evaluations can embarrassingly parallel.

\begin{figure}
\begin{subfigure}{0.33\textwidth}
\hspace*{6pt}
\includegraphics[width=\textwidth, trim={118px 34px 0 32px}, clip]{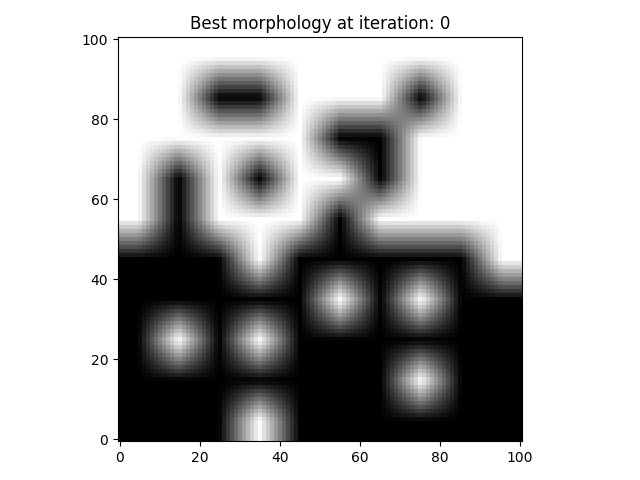}
\caption{Iteration 10}
\end{subfigure}
\begin{subfigure}{0.33\textwidth}
\hspace*{6pt}
\includegraphics[width=\textwidth, trim={118px 34px 0 32px}, clip]{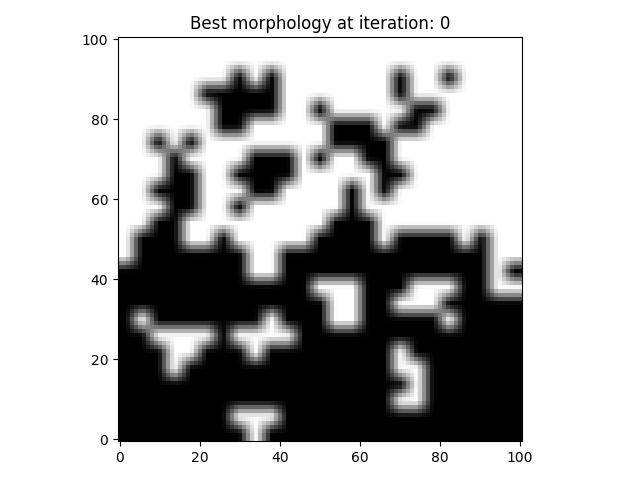}
\caption{Iteration 30}
\end{subfigure}
\begin{subfigure}{0.33\textwidth}
\hspace*{6pt}
\includegraphics[width=\textwidth, trim={118px 34px 0 32px}, clip]{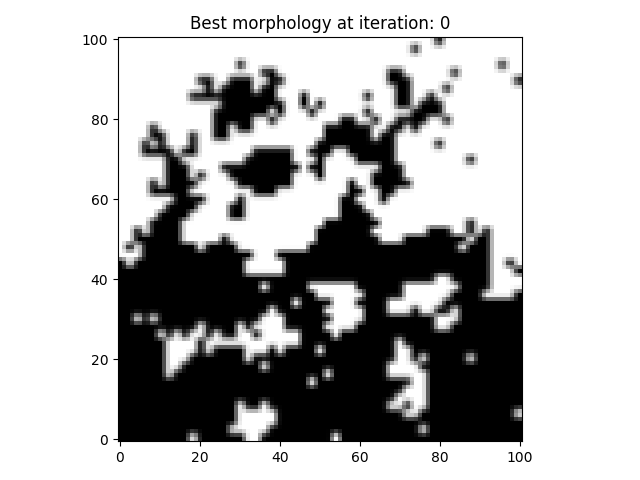}
\caption{Iteration 50}
\end{subfigure}
\caption{Exploration by automated design: The optimization started with a bilayer structure. Notice how the framework directs the formation of finer features }
\label{fig:pbil_design}
\end{figure}

The results of PBIL optimization are very promising. Fig.~\ref{fig:pbil_design} indicates structure at different scales, mimicking finger-like fractal structures. These are very similar to the results presented in~\cite{du2018microstructure}.

\section{Conclusion}
\label{sec:concl}

In this work, we address the issue of designing active layer morphologies to enhance device performance, especially in OPV applications. The usual methods to quantify morphology are either too costly or too simplistic. Hence we take a data-driven approach (DLSP) to create a \emph{morphology quantifier} that can perform fast evaluations. We train a custom designed CNN that reads the morphology and classifies it into $10$ bins of increasing performance metric $J_{sc}$. Using out-of-sample datasets, we confirm that there is no severe over-fitting issues during the training process. Two other standard networks (VGG-16 and ResNet-50) were trained end-to-end independently. It was observed that the custom network, although shallower, gave very similar accuracy. However, our custom network performed much better when visualized using saliency maps as well as when tested on out-of-sample datasets. It identified critical features of the interface in the morphology, which both VGG-16 and ResNet-50 failed to identify consistently. The custom designed network is then used to perform morphology design for achieving enhanced performance. Two approaches were taken to do this -- the first one aims to inform the user about the effect of morphology on performance. The second approach uses the \emph{trust-worthy} network as a fast cost function and performs morphology optimization using PBIL algorithm. It should be noted that this work serves as a proof of concept of using deep neural networks for material morphology quantification. It raises several other interesting questions of how to integrate physical phenomena into the training process. Can these physics based intuitions can be exploited to reduce the demand on the size of data for training? Can a most effective dataset be created to reduce the training data? Can we make the training process more robust to adversarial attacks? All these questions form the scope of future study.

\section{Acknowledgement}
\label{sec:acknow}

This work has been supported in part by the U.S. Air Force Office of Scientific Research under the YIP grant FA9550-17-1-0220. Any opinions, findings and conclusions or recommendations expressed in this publication are those of the authors and do not necessarily reflect the views of the sponsoring agency.

\bibliographystyle{plain}
 \bibliography{main}

\begin{thebibliography}{10}

\bibitem{Alip}
Babak Alipanahi, Andrew Delong, Matthew~T Weirauch, and Brendan~J Frey.
\newblock Predicting the sequence specificities of dna-and rna-binding proteins
  by deep learning.
\newblock {\em Nature biotechnology}, 33(8):831, 2015.

\bibitem{bojarski2016end}
M.~Bojarski, D.~Del~Testa, D.~Dworakowski, B.~Firner, B.~Flepp, P.~Goyal, L.~D.
  Jackel, M.~Monfort, U.~Muller, J.~Zhang, et~al.
\newblock End to end learning for self-driving cars.
\newblock {\em arXiv preprint arXiv:1604.07316}, 2016.

\bibitem{casalegno2010methodological}
M.~Casalegno, G.~Raos, and R.~Po.
\newblock Methodological assessment of kinetic monte carlo simulations of
  organic photovoltaic devices: The treatment of electrostatic interactions.
\newblock {\em The Journal of chemical physics}, 132(9):094705, 2010.

\bibitem{du2018microstructure}
P.~Du, A.~Zebrowski, J.~Zola, B.~Ganapathysubramanian, and O.~Wodo.
\newblock Microstructure design using graphs.
\newblock {\em npj Computational Materials}, 4(1):50, 2018.

\bibitem{Esteva}
A.~Esteva.
\newblock Dermatologist-level classification of skin cancer with deep neural
  networks.
\newblock {\em Nature}, 542(7639):115–118, 2017.

\bibitem{dmref}
National~Science Foundation.
\newblock Designing materials to revolutionize and engineer our future.
\newblock https://www.nsf.gov/funding/pgm\_summ.jsp?pims\_id=505073, 2018.
\newblock Accessed: 2018-11-01.

\bibitem{ghosalhigh}
S.~Ghosal, A.~Akintayo, P.~Boor, and S.~Sarkar.
\newblock High speed video-based health monitoring using 3d deep learning.
\newblock {\em Dynamic Data-Driven Application Systems (DDDAS)}, August 2017.

\bibitem{Ghosal4613}
S.~Ghosal, D.~Blystone, A.~K. Singh, B.~Ganapathysubramanian, A.~Singh, and
  S.~Sarkar.
\newblock An explainable deep machine vision framework for plant stress
  phenotyping.
\newblock {\em Proceedings of the National Academy of Sciences},
  115(18):4613--4618, 2018.

\bibitem{Glorot}
X.~Glorot and Y.~Bengio.
\newblock Understanding the difficulty of training deep feedforward neural
  networks.
\newblock {\em AISTATS}, 2010.

\bibitem{Yosinski}
A.~Nguyen J.~Yosinski, J.~Clune, T.~Fuchs, and H.~Lipson.
\newblock Understanding neural networks through deep visualization.
\newblock {\em arXiv preprint}, 1506.06579, 2015.

\bibitem{ju2017designing}
S.~Ju, T.~Shiga, L.~Feng, Z.~Hou, K.~Tsuda, and J.~Shiomi.
\newblock Designing nanostructures for phonon transport via bayesian
  optimization.
\newblock {\em Physical Review X}, 7(2):021024, 2017.

\bibitem{Simon}
A.~Vedaldi K.~Simonyan and A.~Zisserman.
\newblock Deep inside convolutional networks: Visualising image classification
  models and saliency maps.
\newblock {\em arXiv preprint}, 1312.6034, 2013.

\bibitem{Kingma}
D.~Kingma and J.~Ba.
\newblock Adam: A method for stochastic optimization.
\newblock {\em arXiv preprint}, 1412.6980, 2014.

\bibitem{kodali2012computer}
H.~K. Kodali and B.~Ganapathysubramanian.
\newblock Computer simulation of heterogeneous polymer photovoltaic devices.
\newblock {\em Modelling and Simulation in Materials Science and Engineering},
  20(3):035015, 2012.

\bibitem{marsh2007microscopic}
R.A Marsh, C.~Groves, and N.~C. Greenham.
\newblock A microscopic model for the behavior of nanostructured organic
  photovoltaic devices.
\newblock {\em Journal of applied physics}, 101(8):083509, 2007.

\bibitem{meng2009dynamic}
L.~Meng, Y.~Shang, Q.~Li, Y.~Li, X.~Zhan, Z.~Shuai, R.~GE. Kimber, and A.~B
  Walker.
\newblock Dynamic monte carlo simulation for highly efficient polymer blend
  photovoltaics.
\newblock {\em The Journal of Physical Chemistry B}, 114(1):36--41, 2009.

\bibitem{Mnih}
V.~Mnih.
\newblock Human-level control through deep reinforcement learning.
\newblock {\em Nature}, 518(7540):529–533, 2015.

\bibitem{montavon2017methods}
G.~Montavon, W.~Samek, and K.~M{\"u}ller.
\newblock Methods for interpreting and understanding deep neural networks.
\newblock {\em Digital Signal Processing}, 2017.

\bibitem{ray2012can}
B.~Ray, M.~S. Lundstrom, and M.~A. Alam.
\newblock Can morphology tailoring improve the open circuit voltage of organic
  solar cells?
\newblock {\em Applied Physics Letters}, 100(1):7, 2012.

\bibitem{Silver}
D.~Silver.
\newblock Mastering the game of go with deep neural networks and tree search.
\newblock {\em Nature}, 529(7587):484–489, 2016.

\bibitem{watkins2005dynamical}
P.~K. Watkins, A.~B. Walker, and G.~LB Verschoor.
\newblock Dynamical monte carlo modelling of organic solar cells: The
  dependence of internal quantum efficiency on morphology.
\newblock {\em Nano letters}, 5(9):1814--1818, 2005.

\bibitem{wodo2012graph}
O.~Wodo, S.~Tirthapura, S.~Chaudhary, and B.~Ganapathysubramanian.
\newblock A graph-based formulation for computational characterization of bulk
  heterojunction morphology.
\newblock {\em Organic Electronics}, 13(6):1105--1113, 2012.

\bibitem{wodo2015automated}
O.~Wodo, J.~Zola, B.~S.~S. Pokuri, P.~Du, and B.~Ganapathysubramanian.
\newblock Automated, high throughput exploration of
  process--structure--property relationships using the mapreduce paradigm.
\newblock {\em Materials discovery}, 1:21--28, 2015.

\bibitem{Yamins}
D.L. Yamins and J.J. DiCarlo.
\newblock Using goal-driven deep learning models to understand sensory cortex.
\newblock {\em Nature Neuroscience}, 19(3):356, 2016.

\end{thebibliography}

\end{document}